\documentclass{article} 
\usepackage{iclr2027_conference,times}

\usepackage{amsmath,amsfonts,bm}

\newcommand{\tab}{TAB}

\def\eqref#1{equation~\ref{#1}}

\def\1{\bm{1}}

\DeclareMathAlphabet{\mathsfit}{\encodingdefault}{\sfdefault}{m}{sl}
\SetMathAlphabet{\mathsfit}{bold}{\encodingdefault}{\sfdefault}{bx}{n}

\def\gB{{\mathcal{B}}}

\def\gJ{{\mathcal{J}}}

\def\gM{{\mathcal{M}}}

\DeclareMathOperator*{\argmax}{arg\,max}

\usepackage{microtype}
\usepackage{hyperref}
\usepackage{url}
\usepackage{booktabs}
\usepackage{enumitem}
\usepackage{lineno}

\definecolor{darkblue}{rgb}{0, 0, 0.5}
\hypersetup{colorlinks=true, citecolor=darkblue, linkcolor=darkblue, urlcolor=darkblue}

\usepackage{wrapfig}
\newcommand{\comment}[1]{}
\usepackage{multirow}

\usepackage{url}            
\usepackage{booktabs}       
\usepackage{amsfonts}       
\usepackage{nicefrac}       
\usepackage{microtype}      
\usepackage{amssymb,amsmath, amsthm, bm, cleveref}
\usepackage{algorithm}
\usepackage{algorithmic}
\usepackage[utf8]{inputenc}
\usepackage[T1]{fontenc}
\usepackage{mathtools}
\usepackage{subcaption}
\usepackage{tikz}
\usetikzlibrary{calc}
\usepackage{graphicx}
\usepackage{caption}
\usepackage{bbm}
\usepackage{bm}
\usepackage{tcolorbox}
\usepackage{mdframed}
\usepackage{float}

\title{Not All Turns Are Equally Hard: Adaptive Thinking Budgets For Efficient Multi-Turn Reasoning in Agents}

\author{Neharika Jali, Anupam Nayak, Gauri Joshi \\
Carnegie Mellon University \\
\texttt{njali@cmu.edu}
}

\iclrfinalcopy 
\begin{document}

\maketitle

\begin{abstract}
As LLM reasoning performance plateaus, improving inference-time compute efficiency is crucial to mitigate overthinking and long thinking traces even for simple queries. Prior approaches including length regularization, adaptive routing and difficulty-based budget allocation primarily focus on single-turn settings and fail to address the sequential dependencies inherent in multi-turn reasoning. We formulate multi-turn reasoning as a sequential compute allocation problem and model it as a multi-objective Markov Decision Process. We propose TAB: Turn-Adaptive Budgets, a budget allocation policy  that takes as input the conversation history and learns to maximize task accuracy while respecting global per-problem token constraints, and consequently learns to adaptively allot smaller budgets to easier turns and save an appropriate number of tokens for the crucial harder reasoning steps. We  train and analyze the behavior of \tab~on an auxiliary hierarchical plan-and-execute system for multi-turn math reasoning where solution trajectories are inexpensive to collect. Here, \tab~saves up to $35\%$ tokens and $30\%$ latency over static and off-the-shelf LLM budget baselines. We then transfer the same \tab~policy to environment-grounded agentic workflows spanning sequential tool-use, simulated user-agent dialogue and multi-step terminal tasks, where it generalizes and maintains accuracy while saving up to $20\%$ tokens and $15\%$ latency, without ever being trained on them.
\end{abstract}
 
\section{Introduction} \label{sec:intro}
 
\begin{figure}[htb]
    \centering
    \includegraphics[width=1\linewidth]{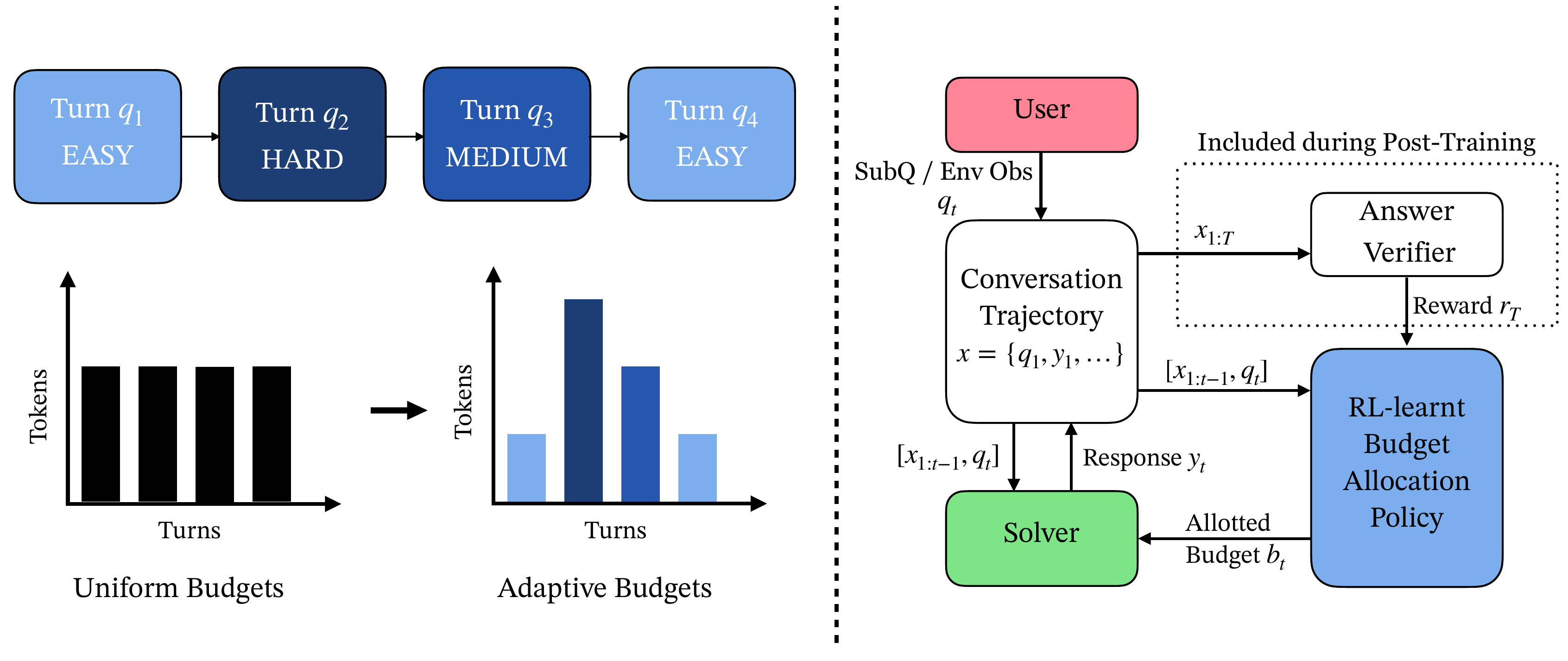}
    \caption{\textit{Left:} A multi-turn reasoning trajectory in agents with turns of varying difficulties, where adaptive difficulty-level based per-turn budget allocations improve compute utilization. \textit{Right:} At every turn $t$ at inference (colored blocks), the user presents an environment observation or sub-question $q_t$, the Budgeter allots a token budget $b_t$ based on $x_{1:t-1}$ and $q_t$, and the Solver generates a response $y_t$. The Budgeter is trained via GRPO with a reward combining accuracy and tokens used.}
    \vspace{-14pt}
    \label{fig:multiTurnSystem}
\end{figure}

As recent gains in LLM reasoning increasingly come from test-time computation \citep{wu2025inference, muennighoff2025s1, snell2024scaling, brown2024monkeys}, a central deployment question is no longer only how to scale inference, but how to spend that inference budget well. While reasoning models are trained to produce long thinking traces before answering \citep{guo2025deepseekr1}, they spend nearly as many tokens on trivial problems as on hard ones, a behavior termed \emph{overthinking} \citep{chen2024overthinking, sui2025stop}. Efficient reasoning matters not only because it reduces cost, but also because it enables more effective test-time scaling under fixed resources, and at the systems level, where under real-world multi-user serving workloads unnecessary reasoning increases latency, memory use and long-context processing costs, reducing throughput \citep{agrawal2024taming, sun2024llumnix, kwon2023efficient, hooper2025squeezed}. Current approaches to efficient test-time reasoning largely follow a common philosophy of first estimating a problem's difficulty, then allocating an appropriate amount of compute, and finally solving the problem under that budget \citep{wang2025make, shen2025dast, zhang2025adaptthink}, thus treating efficient reasoning as a one-shot allocation problem at the instance level.
 
Multi-turn reasoning in agents, often introduced via task or sub-question decomposition to improve performance, makes efficient compute allocation even more important, since a task is not a single query but a sequence of sub-questions, tool calls and environment observations \citep{yao2023react, jimenez2024swebench}. In single-turn settings wasted compute is local, and an unnecessarily long trace only slows and increases the cost of one response, whereas in multi-turn settings inefficiency compounds over time \citep{jeong2025accelerating, gao2024cost} in two ways. First, tokens generated early become part of later context, raising serving costs as the interaction proceeds and diluting the context later steps must reason over \citep{agrawal2024taming, qin2025mooncake, zhang2023h2o, liu2024lost}. Second, under a fixed per-problem budget, tokens spent on an easy turn are no longer available for harder turns later in the trajectory. Consequently, overthinking has been observed to be more damaging for agents than for single-turn solvers \citep{cuadron2025danger}. This is especially problematic in decomposed pipelines, where some sub-questions are routine while others are decisive \citep{zeng2025reinforcing}. Yet multi-turn budget allocation remains relatively underexplored.
 
In this work, we cast multi-turn reasoning as a distinct sequential compute allocation problem. This setting is challenging for several reasons. First, it is characterized by strong \emph{temporal dependency} across turns. A budget decision at the current turn affects not only the immediate response quality, but also the remaining compute, the downstream reasoning trajectory, and the final task outcome, all without visibility into future turns. Second, feedback is delayed. Since rewards are observed, or correctness is verified, only at the end of the trajectory, the problem involves difficult \emph{credit assignment} across turn-level decisions. Third, the objective is inherently \emph{multi-objective}. A successful approach must jointly optimize task accuracy and token efficiency over entire reasoning trajectories. Finally, learning such a policy requires a task reward for every sampled trajectory, which is inexpensive to obtain in verifiable domains such as math reasoning but considerably more \emph{expensive to collect} in the environment-grounded agentic settings we ultimately target, where a single rollout involves tool calls, simulated users and stateful environments. Together, these properties make multi-turn budget allocation fundamentally different from single-turn budget selection and render existing single-turn efficiency methods non-trivial to extend to this setting.

\paragraph{Main Contributions.} Our contributions toward efficient multi-turn reasoning are threefold.
\begin{enumerate}[leftmargin=*]
    \item \textbf{Sequential Budget Allocation Framework.} We formulate multi-turn reasoning as a sequential compute allocation problem formalized as a
    multi-objective Markov Decision Process, and propose \tab: Turn-Adaptive Budgets, a budget allocation
    policy that maximizes task accuracy while respecting
    global per-problem token constraints.
    \item \textbf{Token and Latency efficient Multi-Turn Reasoning}. We  \tab~on an auxiliary
    hierarchical plan-and-execute system for multi-turn math reasoning, where examples are inexpensive to
    collect, and demonstrate a superior accuracy-cost tradeoff, saving up to $35\%$ tokens and $30\%$ latency across five math reasoning benchmarks. This allows us to characterize the learned budget allocation policy in isolation since every policy is evaluated on an identical sequence of turns.
    \item \textbf{Zero-Shot Generalization to Agentic Workflows.} \tab~policy trained on multi-turn math reasoning generalizes to our target environment-grounded agentic
    workflows without incurring the expense of training on them on expensive multi-turn agentic trajectories, saving up to $20\%$ tokens and
    $15\%$ latency while maintaining accuracy on BFCL, Tau-Bench and Terminal-Bench. 
\end{enumerate}
 
\section{Related Works} \label{sec:relatedWorks}
 
\textbf{Inference Time Budget Adaptation Methods.} Inference-time interventions are increasingly used to improve reasoning efficiency by allocating computation more selectively, an idea predating reasoning models in early-exit schemes \citep{schuster2022confident}. One common strategy is model routing, which directs easier queries to smaller or specialized LLMs while reserving larger models for more difficult instances \citep{ding2024hybrid,zhang2025routerr,ong2025routellm,patel2025proxrouter}. Related work has also explored token-level routing, which enables finer-grained compute allocation within a single generation rather than only at the query level \citep{kapoor2026trim}. Another line of work focuses on estimating question difficulty before assigning a reasoning budget \citep{wang2025make,shen2025dast,huang2025adactrl}, while other approaches dynamically switch between thinking and non-thinking modes based on estimated difficulty \citep{zhang2025adaptthink}. More recently, adaptive budget forcing methods have been proposed, using signals such as confidence or token entropy to decide whether additional reasoning is needed \citep{yuan2025reasoning,li2026efficient}. However, most of these methods are designed for single-turn interactions and do not naturally support planning or budget allocation across multi-turn settings, where reasoning costs must be managed adaptively over the course of an interaction. Further, methods for model routing in multi-turn conversations often incur significant infrastructure overheads of hosting multiple model variants, the prefill and/or decode latency associated with cross-model transitions and cost of extra model invocations.

\textbf{Inference Time Planning.} Complementary to the budget adaptation methods are task decomposition (TD) methods that aim to improve performance by decomposing difficult problems into simpler subproblems that can be solved sequentially. Well before GRPO-style post-training became prevalent, prompting-based TD methods had already shown substantial benefits from explicitly structuring intermediate reasoning \citep{wei2022chain,radhakrishnan2023question,wang2023plan,khot2023decomposed}, alongside parallel strategies that aggregate multiple reasoning paths \citep{wang2023selfconsistency,yao2023tree}. More recent reasoning-oriented work has revisited this paradigm by using a stronger model to generate a decomposition or plan and a smaller model to execute the corresponding subtasks \citep{lin2026plan,xin2025scaling}. This strategy has also been widely adopted in agent-based frameworks \citep{qian2025xroutertrainingcostawarellms,erdogan2025plan}. More recently, hierarchical RL approaches have sought to endow models with subgoal decomposition capabilities through training \citep{wang2025hierarchical,ge2025hierarchical,ren2025deepseek}. Nevertheless, the primary focus of most of these methods has been on improving task performance rather than on enabling more efficient reasoning, and the resulting plan is never used as a signal for how to ration compute, which is precisely the signal \tab~All-SubQ exploits (\Cref{sec:results}).

\textbf{Efficiency in Multi-Turn Agents.} Beyond static task decomposition, efficient reasoning is critical for environment-grounded agents interacting with external tools to solve multi-turn tasks. In such interactive settings, early decisions dictate downstream observations, causing reasoning inefficiencies and context window to compound over time. Multi-turn RL work has largely targeted capability rather than cost \citep{zhou2024archer,wang2025ragen}. To optimize compute, recent methods employ agentic routing, delegating routine steps to smaller models and reserving larger ones for critical boundaries under global budgets \citep{qian2025xroutertrainingcostawarellms, zhang2026budget} or apply turn-level credit assignment to penalize wasteful exploratory loops \citep{zeng2025reinforcing}. However, these approaches rely on macroscopic model-switching, which introduces significant cross-model latency and context-transfer overheads. Consequently, they leave unaddressed the core challenge of dynamically modulating the token-level reasoning budget of a single unified solver across multi-turn trajectories.

\textbf{Training Efficient Reasoners.} Following the observation that GRPO-based post-training tends to increase the average length of model responses as training progresses, a growing body of work has sought to mitigate overthinking in LLMs and improve token-efficient reasoning, especially on easier problems \citep{sui2025stop}. One prominent line of research modifies the training objective by incorporating length penalties or related mechanisms to encourage shorter reasoning traces \citep{arora2025training, aggarwal2025l, ayoub2026learning, liu2026learn, qi2025optimizing}. Other approaches rely on oversampling and training on shorter responses \citep{shrivastava2025sample}, adopt two-stage training procedures \citep{song2025walk, fatemi2025concise}. More recently, several works have also explored training efficient reasoners through self-distillation \citep{zhao2026self, sang2026policy}. The central challenge is to reduce unnecessary verbosity without making the model overly conservative, which can hinder exploration at test time. These works can be seen as developing efficient low-level solvers that may be leveraged as a part of a bigger system at inference time.

\section{\tab: Turn-Adaptive Budgets} \label{sec:algo}
 
In this section, we formulate the efficient multi-turn reasoning problem as a sequential compute allocation problem instead of a single-response length-control problem. We model this as a multi-objective Markov decision process and propose \textbf{\tab: Turn-Adaptive Budgets}, a budget allocation policy learned via the GRPO algorithm with a reward function that is a convex combination of the accuracy and adherence to global token budget.

\subsection{Problem Setting} \label{sec:problemSetting}
 
\paragraph{Notation.} In this work, we study multi-turn reasoning under a global per-problem token budget.  Consider a large language model $f_\theta(\cdot)$ with a set of token budgets $\gB$ that generates response $y$ to query $q$ by respecting the allotted token budget $b \in \gB$ as $y \sim f_\theta(q; b)$ \citep{aggarwal2025l}. A multi-turn problem instance $x$ consists of a sequence of turns, $t \in [T]$, with questions or information $q_t$ presented to the language model and its responses $y_t$ as $x = (q_1, y_1, q_2, y_2, \dots, q_T, y_T)$ where $T$ is the total number of turns in the conversation episode. Note that $q_t$ at every turn could represent a sub-question presented by a user, information about the environment presented to an agent or tool uses and their results.  Denote the conversation up to turn $t$ by $x_{1:t} = (q_1, y_1, \dots, q_t, y_t)$. If at each turn $t$, a budget allocation policy $\pi_{\phi}$, parameterized by $\phi$, selects how much reasoning compute $b_t \in \gB$ to allocate to the solver LLM, the response generated can be represented as $y_{t} \sim f_\theta(x_{1:t-1} , q_t; b_t).$

\paragraph{Need for Sequential Decision Making in Multi-Turn Systems.} Current reasoning models are prone to overthinking and spending far too many tokens on easy turns of a multi-turn trajectory in agents leaving little budget for later harder turns. Unlike for single-turn tasks, as demonstrated in \Cref{fig:mainResults,fig:agenticTokens}, static and context-unaware individual turn $q_t$ difficulty based token budget policies are ineffective in the multi-turn setting which presents the following set of challenges. First is the \emph{temporal dependency} and its inherently sequential nature where compute allocation decisions at the current turn made without seeing the future turns or sub-questions determine the future resources remaining, downstream reasoning trajectory and eventual problem accuracy in addition to the quality of the immediate response. Second is the challenge of delayed feedback where rewards presented or accuracy verifiable only at the end of the trajectory resulting in the \emph{credit-assignment} problem making it difficult to discern which specific turn-level budget decisions contributed to the final outcome. Third, the problem requires a \emph{multi-objective optimization} algorithm that can optimize over sequential trajectories to effectively learn the accuracy-tokens tradeoff. Finally, learning such a policy needs a task reward for every sampled trajectory, which is cheap to obtain in verifiable domains such as math reasoning but \emph{expensive to collect} in environment-grounded agentic settings where a single rollout involves tool calls, simulated users and stateful environments.

\subsection{Multi-Turn Reasoning in Agents as a Markov Decision Process} \label{sec:MDP}
 
We frame the sequential allocation of token budgets as a scalarized multi-objective MDP as follows \citep{roijers2013survey, van2014multi}. Consider a Markov decision process $\gM$. The state at turn $t$ represents the problem trajectory so far $s_t = [x_{1:t-1}, q_t]$. The action represents the token budget chosen $b_t \sim \pi_\phi(\cdot | x_{1:t-1}, q_t)$. The transition to the next state consists of concatenating the response generated by the solver LLM $y_t$ to the problem trajectory as $x_{1:t} = (q_1, y_1, \dots, q_t, y_t)$. Consider tasks with verifiable outcomes and represent accuracy of the generated solution as $\text{acc}(x) = \mathbb{1}\{\texttt{verify}(x)\}$ where $\texttt{verify}(\cdot)$ is the success criterion supplied by the benchmark. For the math reasoning benchmarks this is an exact match of the final response against the ground truth label $y_T^\star$, and for the environment-grounded benchmarks, on which the policy is evaluated but never trained, it is the corresponding state-based or tool-trace criterion (\Cref{app:implementation}). The objective of the MDP, thus, is to find an optimal budget allocation policy $\pi^\star_\phi$ that maximizes the expected accuracy of the reasoning task subject to a global token budget $B$ modelled as a soft penalty as
\begin{equation} \label{eq:MDP}
    \pi^\star_\phi = \argmax_\pi \mathbb{E}_{x, \pi} \left[\text{acc}(x) - \lambda \max\left(0, \sum_{t=1}^T b_t - B\right)\right],
\end{equation}
where the budget violation penalty weight $\lambda$ characterizes the trade-off between accuracy and token efficiency. Alternative reward scalarizations can also be designed to promote concise reasoning, including exact-target, maximum-length, and smoother sigmoid-based penalties \citep{aggarwal2025l, arora2025training, ayoub2026learning}. We use a hinge penalty because our goal is not to match a target length, but to discourage violating a global per-problem budget. This formulation aligns closely with the deployment objective and is less susceptible than naive additive penalties to pathological under-allocation.

\subsection{Training Turn-Adaptive Budget Allocation Policy via GRPO} \label{sec:GRPO}
\paragraph{Reward Function.} We learn the optimal budget allocation policy $\pi^\star_\phi(b_t | x_{1:t-1}, q_t)$ where $b_t \in \gB$ using the Group Relative Policy Optimization (GRPO) algorithm \citep{shao2024deepseekmath} as follows. Note that we adopt GRPO over Proximal Policy Optimization (PPO) \citep{schulman2017proximal} as it eliminates the need for a learned value function, which is notoriously difficult to estimate accurately in long-horizon, sparse-reward settings such as multi-turn reasoning. Now, for each training problem $i$, we sample a group of $G$ multi-turn trajectories $\{x^{(i, g)}\}_{g=1}^G$. Recall that for every turn $t$ of a trajectory, the budget allocator observes the current conversation state, $[x_{1:t-1}^{(i,g)}, q_t^{(i,g)}]$, chooses the budget tokens by sampling $b_t^{(i,g)} \sim \pi_\phi(\cdot | x_{1:t-1}^{(i,g)} , q_t^{(i,g)})$ and supplies it to the LLM, which generates the response $y_t^{(i,g)} \sim f_\theta(x_{1:t-1}^{(i,g)}, q_t^{(i,g)}; b_t^{(i,g)})$. After the final synthesis step, each trajectory receives a single terminal reward $r^{(i,g)}$ that scalarizes correctness and token efficiency. We design the reward function\footnote{Note that while training, we use the number of tokens actually used by the solver LLM at every turn instead of the number of tokens allotted by the budget policy.} to learn the optimal policy (\ref{eq:MDP}) as 
\begin{equation} \label{eq:reward}
    r^{(i,g)}\left(x^{(i,g)}\right) = \text{acc}\left(x^{(i,g)}\right) - \lambda \max\left(0, \sum_{t=1}^T b_t^{(i,g)} - B \right)
\end{equation}
where $\text{acc}(x) \in \{0,1\}$ denotes the correctness of the generated solution, $B$ is the target global token budget and $\lambda$ is a penalty weight.

\paragraph{Advantage Estimation and Policy Optimization.} Following GRPO, rewards are normalized within each group $G$ and advantages estimated as $\hat{A}^{(i,g)}=(r^{(i,g)}-\mu^{(i)})/\sigma^{(i)}$, where $\mu^{(i)}=\mathrm{mean}(\{r^{(i,g)}\}_{g=1}^G)$ and $\sigma^{(i)}=\mathrm{std}(\{r^{(i,g)}\}_{g=1}^G)$.
Note that since the reward is terminal and obtained at the end of the trajectory, the same trajectory-level advantage $\hat{A}^{(i,g)}$ is assigned to all turns $t$ in $x^{(i,g)}$. This may appear to be in tension with the credit-assignment challenge of \Cref{sec:problemSetting}, since a uniform advantage does not by itself separate an efficient turn from a wasteful one. Note however that no per-turn reward is available to credit an individual allocation, because a budget is only good or bad in light of how the remaining turns unfold, and the group baseline supplies this missing signal instead. Trajectories in a group share a problem but differ in their sampled allocation patterns, so normalizing against the group mean identifies which \emph{allocation profiles} succeed under the global budget rather than which single turns do. Finer-grained turn-level advantages remain an interesting direction for future work. We then maximize a clipped policy objective over all turn-level actions, without a value network as
\begin{gather}
    \gJ_{GRPO}(\phi) = \mathbb{E}_{x^{(i, g)}}\left[\frac{1}{|b_t^{(i,g)}|} \sum_{k=1}^{|b_t^{(i,g)}|}\min\left(\rho^{(i,g)}_{t, <k} \hat{A}^{(i,g)}, \text{clip}\left(\rho^{(i,g)}_{t, <k}, 1-\epsilon, 1+\epsilon \right)\hat{A}^{(i,g)} \right) \right] \\
    \text{where} \quad\quad \rho^{(i,g)}_{t, <k}  = \frac{\pi_\phi\left(b^{(i,g)}_{t, <k} | x^{(i,g)}_{1:t-1} , q_t\right)}{\pi_{\phi_{old}}\left(b^{(i,g)}_{t, <k} | x^{(i,g)}_{1:t-1}, q_t \right)}, \nonumber
\end{gather}
and $\pi_\phi$, $\pi_{\phi_{old}}$ refer to the policy in the current and previous step of training respectively. Optionally, we add a reference-policy regularizer to limit policy drift with $\beta\ge 0$, leading to the composite loss function $$\mathcal{J}(\phi)=\mathcal{J}_{\mathrm{GRPO}}(\phi)-\beta\,D_{\mathrm{KL}}(\pi_\phi || \pi_{ref})$$.

\subsection{Inference in Multi-Turn Reasoning Systems with \tab} \label{sec:InferenceTAB}
As demonstrated in \Cref{fig:multiTurnSystem}, a problem presented to the multi-turn reasoning system at inference time is processed as follows. At every turn $t \in [T]$, a question or information $q_t$ along with the trajectory thus far $x_{1:t-1}$ is presented to the solver LLM. The \tab~budget allocation policy takes as input the conversation trajectory thus-far and the current sub-question and allots a token budget as $b_t \sim \pi_\phi(\cdot | x_{1:t-1}, q_t)$. The solver LLM then generates a response as $y_t \sim f_\theta(x_{1:t-1}, q_t; b_t)$.

\section{Experiments} \label{sec:experiments}
 
\subsection{Setup} \label{sec:expSetup}
 
\paragraph{Datasets.} The focus of our work is environment-grounded workflows where agents interact with dynamic external feedback, for which we use the multi-turn subset of the Berkeley Function Calling Leaderboard v4 \citep{patil2025bfcl} for sequential tool-use, the retail and banking subsets of Tau-Bench \citep{barres2025tau2bench} for user-agent dialogues, and the medium-difficulty tasks of Terminal-Bench 1.0 \citep{merrill2026terminalbench} for multi-step terminal and coding workflows. Since trajectories in these workflows are expensive to collect, we train instead on an auxiliary hierarchical plan-and-execute system for multi-turn math reasoning, and evaluate it on the test splits of MATH-500 \citep{lightman2024lets}, AMC23 \citep{mathai_amc23}, MATH Level-5 \citep{hendrycks20221measuring}, OlympiadBench \citep{he2024olympiadbench}, and AIME25 \citep{aime25}. Note that the Budgeter never sees a trajectory from BFCL, Tau-Bench or Terminal-Bench while training.

\paragraph{Multi-Turn Agentic Systems and Models.} Our system consists of three LLMs interacting as demonstrated in \Cref{fig:multiTurnSystem} in the ReAct framework \citep{yao2023react}. (a) \emph{User}: an LLM providing the sequential context $q_t^{(i)}$ for the turn, acting as the environment returning tool-call and code-run observations or as the human user in the environment-grounded systems, and as a Planner decomposing a complex problem into a static sequence of sub-questions in the hierarchical system. (b) \emph{Budgeter}: the budget allocation policy LLM $\pi_\phi$ outputting a token budget for every turn as $b_t^{(i)} \sim \pi_\phi(\cdot | x_{1:t-1}^{(i)}, q_t^{(i)})$. (c) \emph{Solver}: the solver LLM $f_\theta$ generating the response as $y_t^{(i)} \sim f_\theta(x_{1:t-1}^{(i)}, q_t^{(i)}; b_t^{(i)})$.
 
We use Qwen3-4B as the Budgeter for both the auxiliary multi-turn math reasoning system and the environment-grounded workflows. For BFCL and the math benchmarks we use Qwen3-8B as the User and L1-Qwen3-8B-Exact \citep{aggarwal2025l} as the Solver, for Tau-Bench GPT-5.4-nano for both roles, and for Terminal-Bench GPT-5.4-nano as the Solver. Note L1-Qwen3-8B-Exact is an open-source model RL-trained to adhere to user-specified length constraints, which we choose as our Solver as it enables control over the number of response tokens generated from a discrete set.\footnote{While a higher allocated token limit pushes the model towards longer reasoning, the allocated budget may not always be used in its entirety by the generated response.} 
 
We consider budget sets $\gB \in \{256, 512, 1024, 2048, 4096\}$ for L1-Qwen3-8B-Exact and $\gB \in \{\text{none, low, medium, high}\}$ for GPT-5.4-nano, and note that the budget control mechanism can be swapped modularly. The Budgeter emits a difficulty level in $\{0,\dots,4\}$ rather than a raw token count, mapped in order onto the L1-Qwen3-8B-Exact budgets as $0 \mapsto 256, \dots, 4 \mapsto 4096$, and onto the GPT-5.4-nano settings as $\{0,1\} \mapsto \text{none}$, $2 \mapsto \text{low}$, $3 \mapsto \text{medium}$, $4 \mapsto \text{high}$. All GPT-5.4-nano roles run through the API with budgets enforced via the reasoning-effort parameter, while Qwen models are served locally on NVIDIA H100 GPUs (\Cref{app:implementation}). Note that since the Budgeter allots a categorical difficulty level rather than a raw token count, it learns how demanding a turn is relative to the rest of its trajectory instead of how many tokens a particular Solver consumes. We find that this signal survives both a change of Solver and of domain.

\paragraph{Baselines.} We compare our method, \tab, against the following baselines. (i) \emph{Static}: every turn gets the same budget $b_t = b \in \gB$ $\forall t \in [T]$, from the budget set corresponding to the Solver in use. (ii) \emph{LLM-Judge Individual}: an off-the-shelf LLM allocator with the individual current turn as input, $b_t \sim \pi_{\phi}(\cdot | q_t)$. (iii) \emph{LLM-Judge Multi-Turn}: an off-the-shelf LLM allocator with the conversation trajectory so far and the current sub-question as input, $b_t \sim \pi_{\phi}(\cdot | x_{1:t-1}, q_t)$. Note that in both baselines (ii) and (iii), the off-the-shelf LLM is not finetuned for budget allocation. For a controlled comparison, all LLM-Judge baselines use the same base model as the corresponding Budgeter, prompted to emit a difficulty level and parsed through the same mapping.

\begin{figure}[t]
    \centering
    \includegraphics[width=0.9\linewidth]{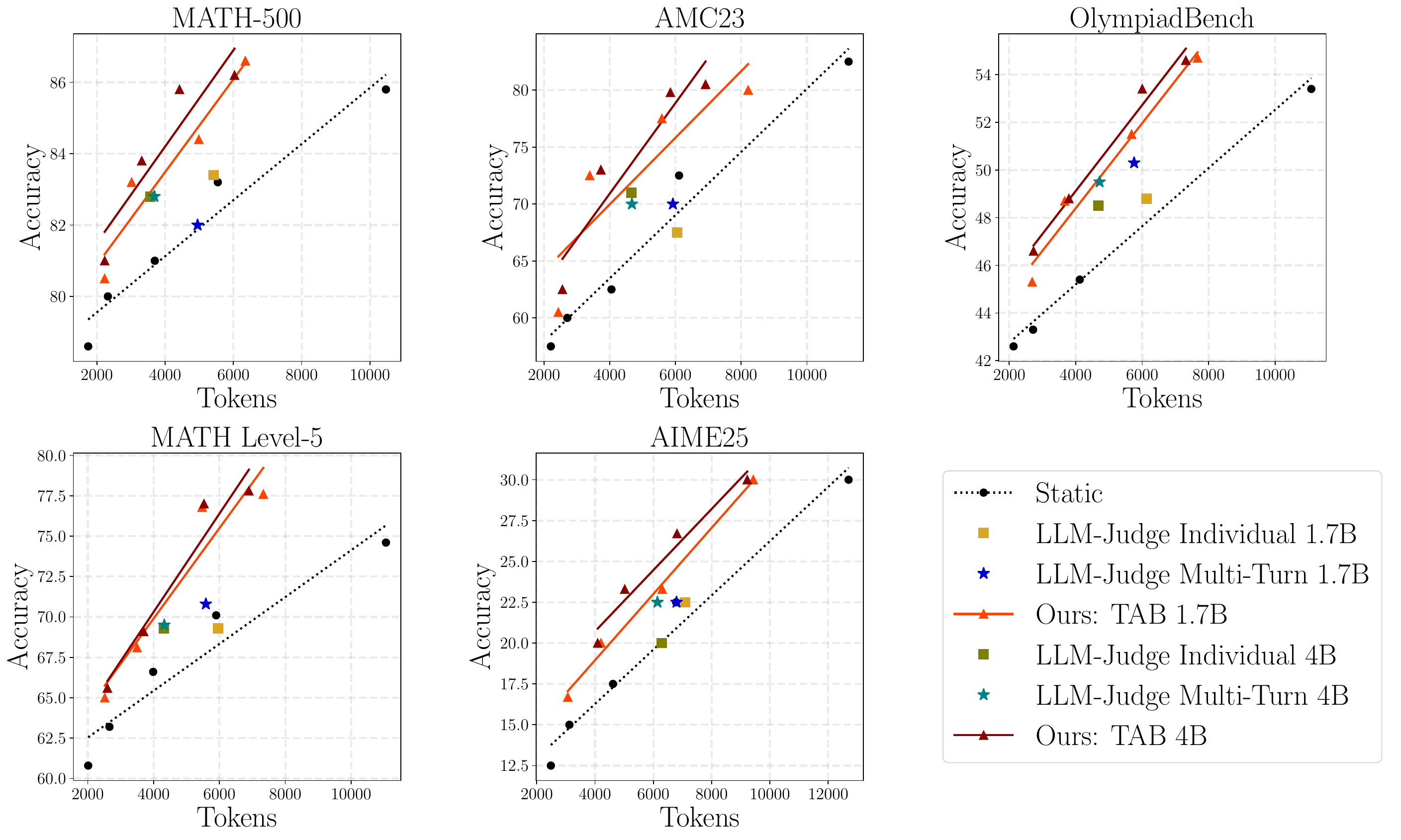}
    \vspace{-10pt}
    \caption{Accuracy-tokens tradeoff of \tab~learned by training Qwen3-4B and Qwen3-1.7B budget policies with budget penalties $B \in \{3k, 5k, 8k, 10k\}$ that saves up to $35\%$ total tokens (User + Budgeter + Solver tokens) over baselines across 5 math reasoning datasets with a L1-Qwen3-8B-Exact Solver. The Qwen3-4B Budgeter matches or improves upon Qwen3-1.7B at nearly every allocation, and it is this policy that is carried unchanged to the agentic benchmarks of \Cref{fig:agenticTokens}.}
    \vspace{-10pt}
    \label{fig:mainResults}
\end{figure}

\begin{wrapfigure}{r}{0.40\linewidth}
    \centering
    \vspace{-5pt}
    \includegraphics[width=1\linewidth]{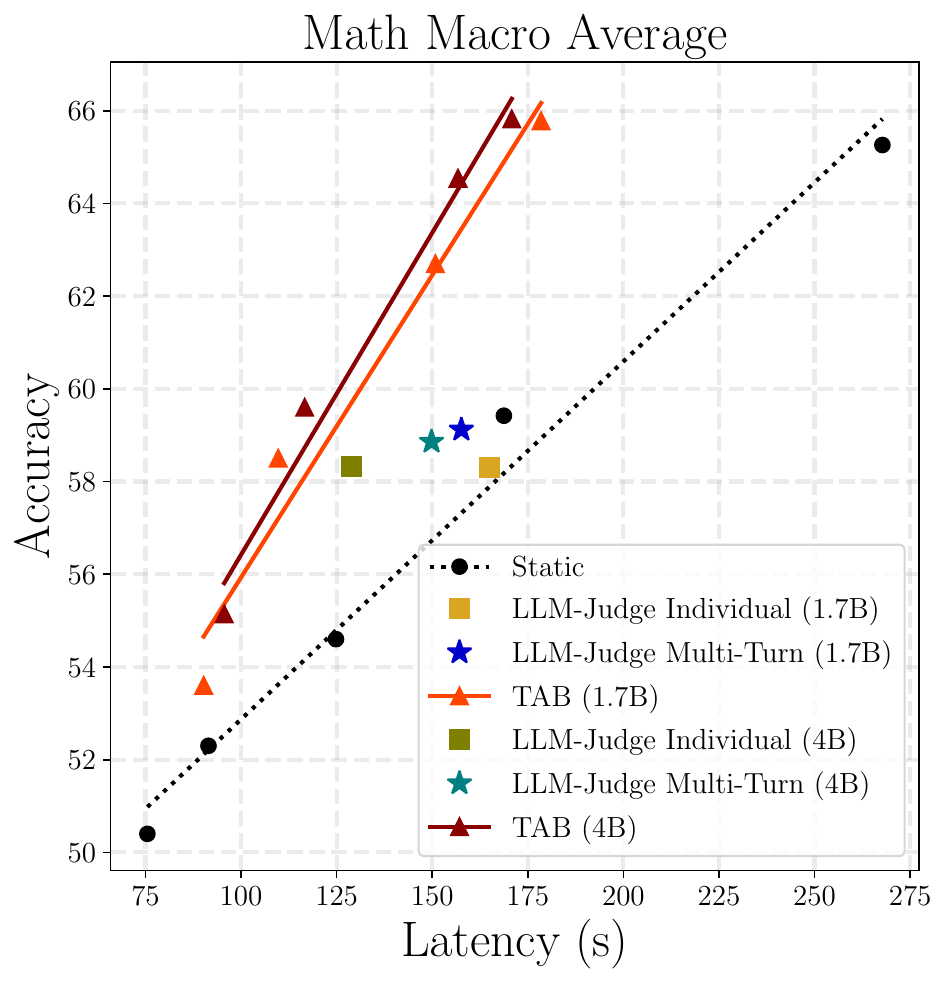}
    \caption{On the auxiliary math task, \tab~reduces overall system latency by up to 30\%, as the time saved by generating fewer Solver tokens more than offsets the Budgeter's overhead. Scaling the Budgeter from 1.7B to 4B also establishes a superior frontier.}
    \label{fig:latency}
    \vspace{-25pt}
\end{wrapfigure}
 
\paragraph{Budget Allocation Policy Training.} As depicted in \Cref{fig:multiTurnSystem}, we RL-train our Qwen3-4B / Qwen3-1.7B Budgeters, $\pi_\phi$, using the GRPO-based algorithm of \Cref{sec:GRPO} with penalty weight $\lambda = 0.001$ in \cref{eq:reward}, on Level 5 questions from the train split of MATH \citep{hendrycks20221measuring}. Note that this is the only data the Budgeter is trained on, and the policies evaluated in \Cref{sec:resultsAgentic} are these very policies applied to the agentic benchmarks without any further finetuning. Instead of full-finetuning we LoRA finetune \citep{hu2022lora}, with all hyperparameters listed in \Cref{app:implementation}. Note that a separate Budgeter policy is trained for each global budget $B \in \{3k, 5k, 8k, 10k\}$.

\subsection{Training on the Auxiliary Task: \\ Multi-Turn Math Reasoning} \label{sec:results}
 
We first train \tab~on a hierarchical plan-and-execute multi-turn system for math reasoning, where the User first decomposes the task into a static sequence of sub-questions $q_1, \dots, q_T$ (a.k.a the plan) and the Budgeter and Solver process this fixed sequence one sub-question at a time (a.k.a the execution). We train on this multi-turn math reasoning system rather than directly on the workflows of \Cref{sec:resultsAgentic} because rollouts here need no tool calls, simulated environments or complex environment resets thus making trajectories inexpensive to collect. Further, the sub-questions here are fixed apriori, so every policy is evaluated on an identical sequence of turns. Token savings thus can be attributed to the policy rather than orchestrator or planning variations and enables us to analyze the characteristics of the learned policy in isolation. Token counts reported here are a sum of the User, Budgeter and Solver tokens.

\begin{figure*}[t]
\centering
\begin{subfigure}{.5\textwidth}
  \centering
  \includegraphics[width=\linewidth]{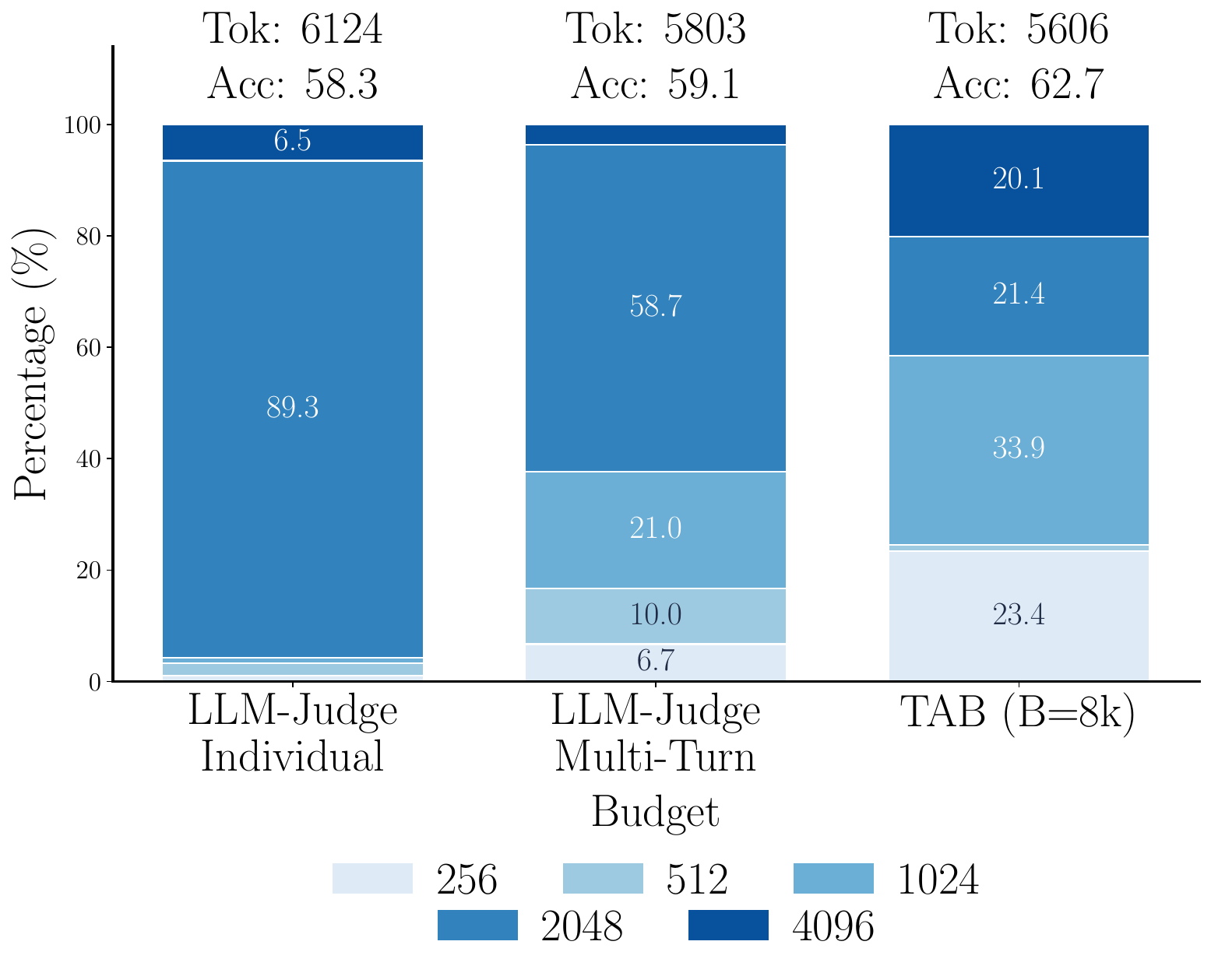}
  \vspace{-10pt}
  \caption{}
  \label{fig:histogramsResults}
\end{subfigure}%
\begin{subfigure}{.5\textwidth}
  \centering
  \includegraphics[width=\linewidth]{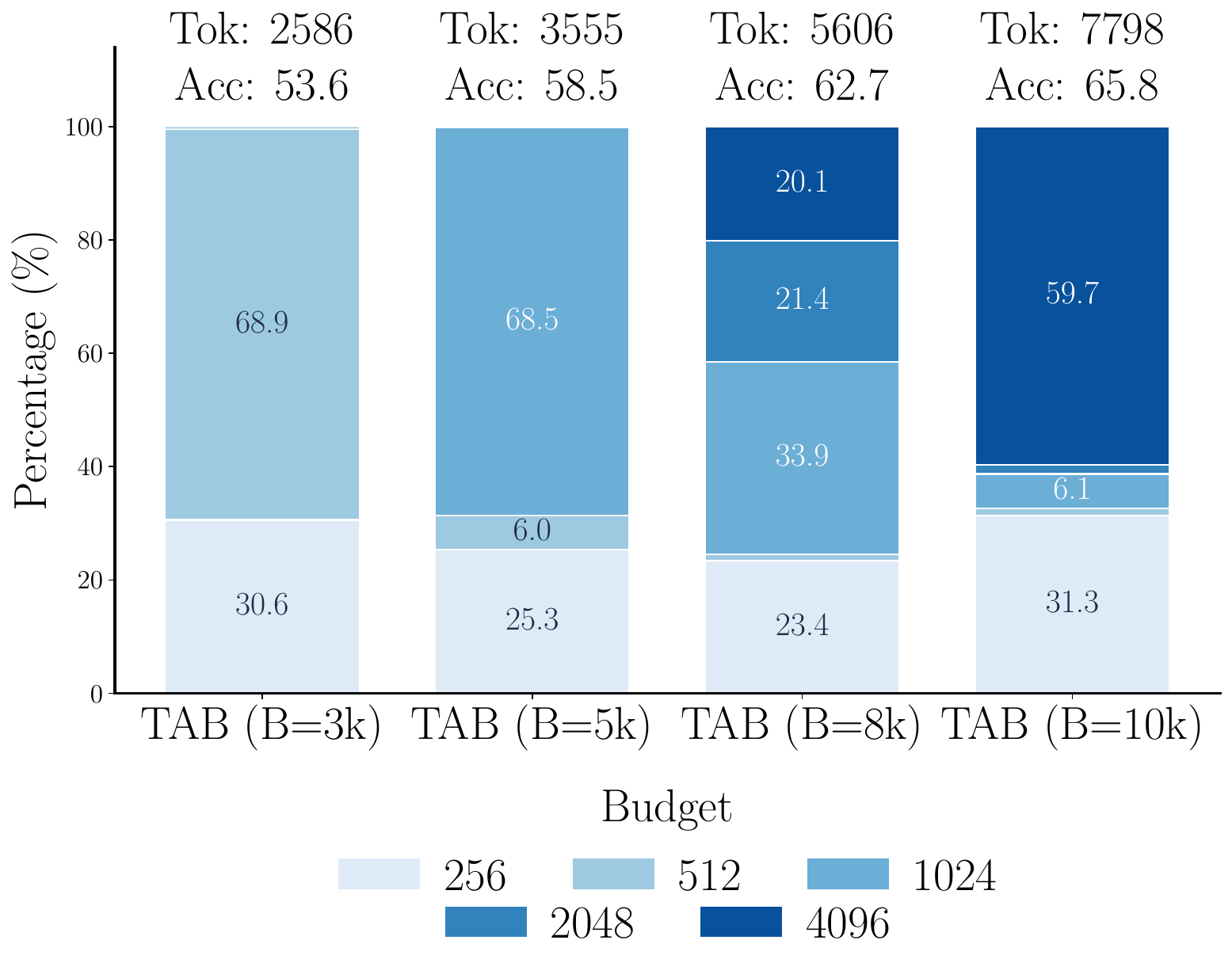}
  \vspace{-10pt}
  \caption{}
  \label{fig:histogramsPareto}
\end{subfigure}%
\caption{Composition-bar plots of allotted tokens per-turn aggregated over all Math reasoning datasets with the accuracy and total tokens used corresponding to each policy listed above their bars. On the left, \tab~learns to adaptively allocate tokens better leading to $4.4$ percentage-point higher accuracy while saving $8.5\%$ total tokens. On the right, are \tab~policies trained with budget penalties $B \in \{3k, 5k, 8k, 10k\}$ and as per-problem global budget $B$ increases, the learned policy shifts towards allotting higher per-turn budgets.}
\vspace{-10pt}
\label{fig:compositionBars}
\end{figure*}

\textbf{Accuracy, Tokens and Allocation Behavior.} As demonstrated in \Cref{fig:mainResults} and \Cref{fig:latency}, \tab~saves up to $35\%$ tokens while maintaining or improving upon baseline accuracy with a corresponding $30\%$ reduction in time taken. Note that \tab~sits between $1.2$ and $5.0$ percentage points above the better of the two LLM-Judge baselines at comparable token counts on every dataset. \Cref{fig:compositionBars} shows where these savings come from. Unlike the LLM-Judge baselines, which cluster around high-compute allocations such as 2048 tokens per turn, \tab~scales down budgets for the easier intermediate steps while reserving tokens for the turns requiring complex reasoning, and retains this pattern even as $B$ grows. Further qualitative examples are presented in \Cref{app:examples}. Ablating the Budgeter size in \Cref{fig:mainResults} shows a gain of up to $2.3$ percentage points at matched tokens from scaling Qwen3-1.7B to Qwen3-4B, as stronger models produce more refined difficulty estimates. We further demonstrate in \Cref{app:ood} that \tab~also improves the performance-efficiency tradeoff on general reasoning tasks that are out-of-domain for math.
 
\textbf{\tab~All-SubQ: Planning and Value of Future Context} A hierarchical system commits to the plan $q_1, \dots, q_T$ before any budget is allotted, so the future sub-questions are available at turn $1$ and \tab~as described thus-far does not use them. We therefore propose \tab~All-SubQ, which allots tokens as $b_t \sim \pi_\phi(\cdot | x_{1:t-1}, q_1, \dots, q_T)$. As demonstrated in \Cref{fig:allSubQResults} in \Cref{app:allsubq}, \tab~All-SubQ saves up to $40\%$ tokens over baselines without loss in accuracy. Since lookahead is the only difference between the two variants, this highlights that explicit knowledge of future steps allows for more precise compute allocation.

\subsection{\tab~on Environment-Grounded Agentic Workflows} \label{sec:resultsAgentic}
 
\begin{figure}[t]
    \centering
    \includegraphics[width=0.95\linewidth]{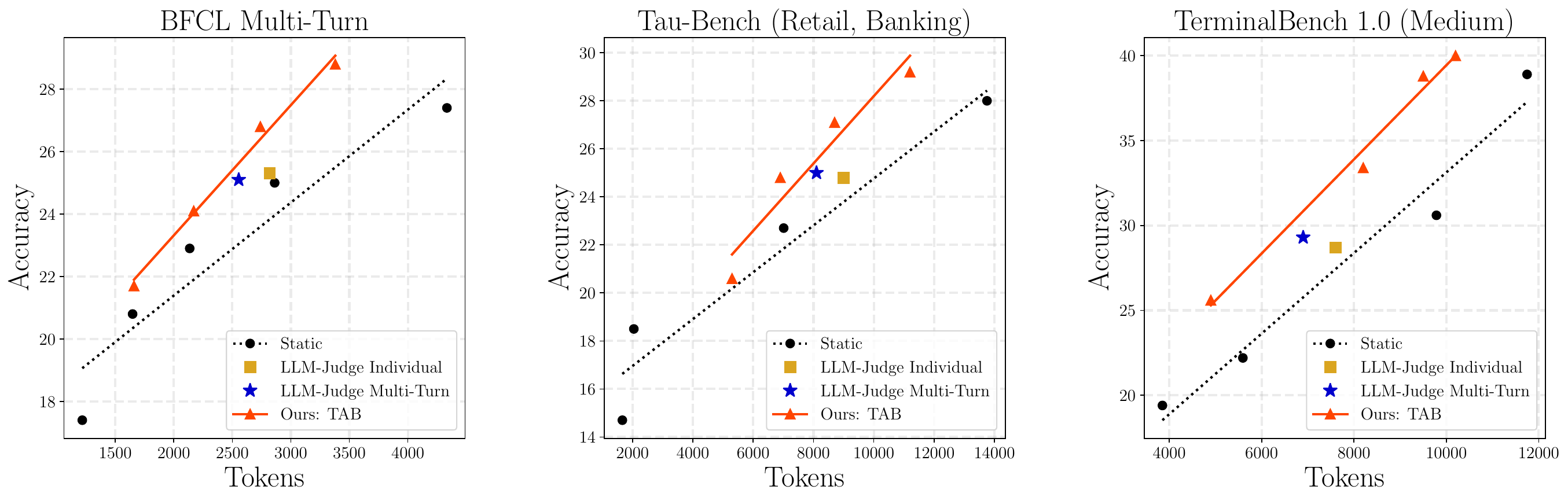}
    \vspace{-5pt}
    \caption{Accuracy-Tokens tradeoff of our RL-learned budget allocation policy \tab~formed by training Qwen3-4B on MATH dataset with budget penalties $B \in \{3k, 5k, 8k, 10k\}$ saves up to $20\%$ total tokens (Budgeter + Solver tokens) over baselines. BFCL uses L1-Qwen3-8B-Exact as the Solver, and Tau-Bench and Terminal-Bench use GPT-5.4-nano as the Solver.}
    \label{fig:agenticTokens}
    \vspace{-5pt}
\end{figure}

We now turn to the environment-grounded agentic workflows that are the focus of our work, namely BFCLv4 Multi-Turn, Tau-Bench and Terminal-Bench, where every $q_t$ is an observation returned by the environment in response to the previous action and the turns are not known ahead of time. We evaluate these benchmarks using the \tab~policies trained on multi-turn math reasoning (\Cref{sec:results}) with the mapping from the emitted difficulty level onto the Solver's budget control mechanism adjusted appropriately. Token counts reported here comprise the Budgeter and Solver tokens. As shown in \Cref{fig:agenticTokens}, \tab~consistently defines a superior accuracy-compute tradeoff, saving up to $20\%$ tokens over baselines while improving or maintaining accuracy. On BFCL, \tab~reaches $28.8\%$ accuracy using $3.4$k tokens while the best Static allocation reaches only $27.4\%$ using $4.3$k. The pattern holds on Tau-Bench, where \tab~attains $29.2\%$ with $11.3$k tokens against Static's $28.0\%$ with $13.8$k, and on Terminal-Bench, $38.8\%$ with $9.5$k against $38.9\%$ with $11.7$k.
 
\begin{figure}[t]
    \centering
    \includegraphics[width=0.95\linewidth]{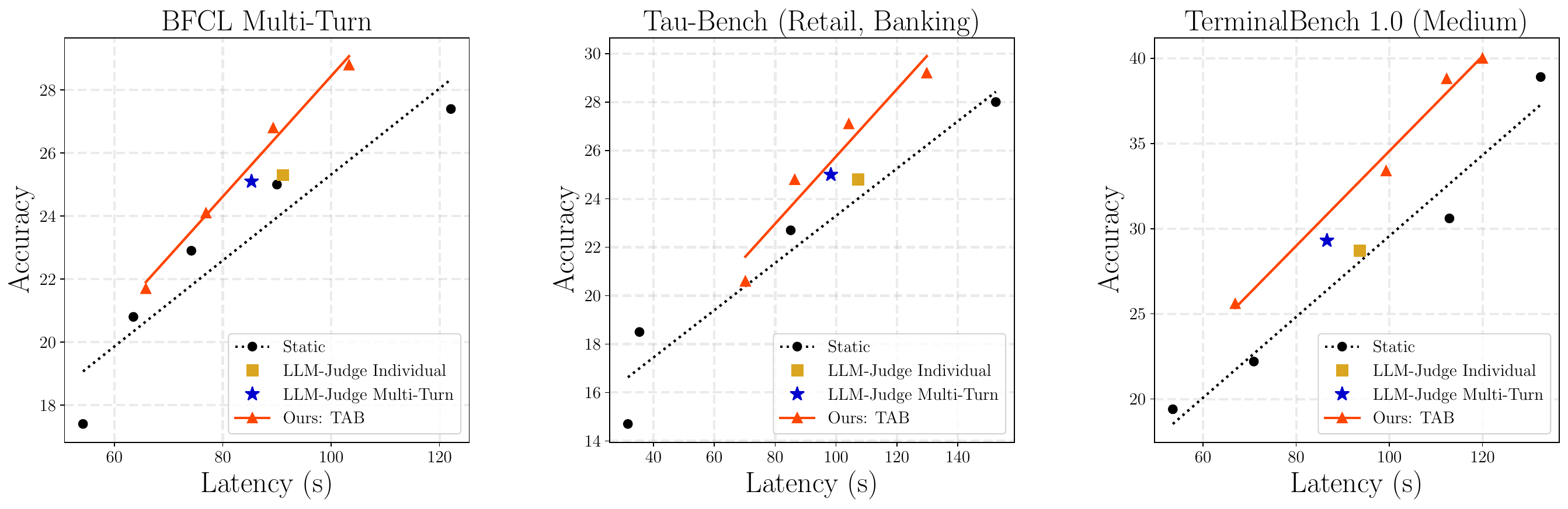}
    \vspace{-5pt}
    \caption{Accuracy-Latency tradeoff of \tab~on the environment-grounded agentic workflows. Even after accounting for the inference latency overhead of querying the Budgeter for turn-level allocations, \tab~saves up to $15\%$ in overall system latency by significantly reducing the number of Solver output tokens. Note that BFCL uses the L1-Qwen3-8B-Exact Solver hosted locally while the other two benchmarks use the GPT-5.4-nano API, with the Budgeter Qwen3-4B hosted locally in all cases.}
    \label{fig:agenticLatency}
    \vspace{-12pt}
\end{figure}
 
Comparing the two LLM-Judge baselines isolates how much of this gain comes from the conversation history rather than the learned policy. Across the three benchmarks the two judges differ by at most $0.6$ percentage points, whereas at comparable budgets \tab~is $1.5$, $2.3$ and $4.7$ points above the individual judge on BFCL, Tau-Bench and Terminal-Bench respectively. On Tau-Bench \tab~matches the accuracy of the individual judge exactly, at $24.8\%$, using $6.9$k tokens instead of $9.0$k. Access to the trajectory is therefore worth far less than learning how to spend against a global budget. Gains are smallest at the tightest budget $B=3k$, where few budget buckets remain for adaptive allocation, and grow with $B$. Finally, \Cref{fig:agenticLatency} shows the reduction in Solver tokens more than offsets the overhead of querying the Budgeter at every turn, yielding up to a $15\%$ reduction in end-to-end latency at matched accuracy across the three benchmarks.


\section{Conclusion}
 
We formulate efficient multi-turn reasoning as a sequential compute allocation problem. By formalizing this as a scalarized multi-objective MDP, we introduced \tab, a Turn-Adaptive Budgeting policy that optimizes the accuracy-efficiency tradeoff over entire trajectories. We trained \tab~on an auxiliary hierarchical plan-and-execute system for multi-turn math reasoning, where it saves up to $35\%$ tokens and $30\%$ latency, and the same policy generalizes to agentic workflows without any further finetuning saving up to $20\%$ tokens and $15\%$ latency. This allows us to avoid the computational expense of training on sampled multi-turn agentic trajectories. Our experiments show that the gains arise from three factors: redistributing compute across turns rather than uniformly compressing reasoning, training the allocation policy rather than relying on an off-the-shelf allocator with the same conversation context, and learning trajectory-level structure that transfers across domains. This enables \tab~to train on an inexpensive auxiliary task while generalizing to agentic environments.

 
\newpage
\subsection*{AI use statement}
 
 
In this work, we used generative AI tools for polishing our writing and writing code.
We have not used generative AI tools for formulating our problem, developing algorithms or designing experiments and the rest of the required disclosure tasks are not applicable to this work.
We have reviewed all AI-assisted work including verifying and testing LLM-generated text and code for correctness. We take responsibility for the final content of this work, including text, claims or artifacts produced with the aid of generative AI.

\subsection*{Ethics statement}
This work studies how to improve the compute efficiency of multi-turn reasoning and agentic systems through adaptive allocation of reasoning budgets. We do not identify any new application-specific ethical risks arising from the proposed budget allocation method beyond those already posed by existing language models and agentic systems. The authors have read the ICLR Code of Ethics and believe that the paper complies with it.

\subsection*{Reproducibility statement}
We provide the training hyperparameters in the appendix and the anonymized training code in a ZIP file submitted as supplementary material. We plan to release all assets including the code and model checkpoints upon publication.

\bibliography{references}
\bibliographystyle{iclr2027_conference}
 
\newpage
\appendix
\section*{Appendix}

\section{Implementation Details} \label{app:implementation}
 
\paragraph{Task Success Criteria.} As noted in \Cref{sec:MDP}, our formulation requires only that the terminal success indicator $\text{acc}(x) = \mathbb{I}\{\texttt{verify}(x)\}$ be computable at the end of a trajectory, and we use the success criterion supplied by each benchmark. For the hierarchical math reasoning benchmarks this is an mathematical match of the final answer against the ground truth label. For BFCL multi-turn it combines a state-based check, which compares the backend state after execution against the expected final state while excluding private attributes, with a response-based check, which requires the model's sequence of function calls to contain the ground-truth minimal call path as a subset rather than to match it exactly. For Tau-Bench the reward of a task is the product of the checks listed in its reward basis, drawn from a database check that compares the final environment state against the state obtained by replaying the reference actions, a check on the information the agent is required to communicate to the user, action matching against the annotated solution calls, and environment or natural-language assertions on the final state. The retail domain is scored on the database and communication checks, while the knowledge-retrieval banking domain, where the agent answers from unstructured documents rather than mutating a database, is scored on the criteria its task definitions specify. For Terminal-Bench it is the outcome of the test script shipped with each task.

\paragraph{Alternative Reward Scalarizations.} The hinge penalty of \cref{eq:MDP} is one of several scalarizations that promote concise reasoning, alongside exact-target, maximum-length and smoother sigmoid-based penalties \citep{aggarwal2025l, arora2025training, ayoub2026learning}. We use a hinge because our objective is adherence to a global per-problem budget rather than agreement with a target length. A naive additive length penalty charges every token equally and therefore pushes uniformly toward under-allocation, including on the turns that determine correctness, whereas the hinge is exactly zero below $B$ and leaves the policy free to redistribute tokens across turns.
 
\paragraph{Budget Action Space and Enforcement.} The Budgeter emits a difficulty level in $\{0,1,2,3,4\}$ rather than a raw token count. For the L1-Qwen3-8B-Exact Solver these levels map to $\gB = \{256, 512, 1024, 2048, 4096\}$ in order, and the selected value is passed to the Solver as its user-specified length constraint. For the GPT-5.4-nano Solver the levels are mapped as $\{0,1\} \mapsto \text{none}$, $2 \mapsto \text{low}$, $3 \mapsto \text{medium}$ and $4 \mapsto \text{high}$, and are enforced through the reasoning-effort input parameter of the API request rather than through instructions placed in the system prompt. Because the learned policy acts on levels rather than on token counts, the same policy can be evaluated against either Solver without retraining. When a generated string cannot be parsed into a level, we default to the medium budget.
 
\begin{figure}[H]
    \centering
    \includegraphics[width=1\linewidth]{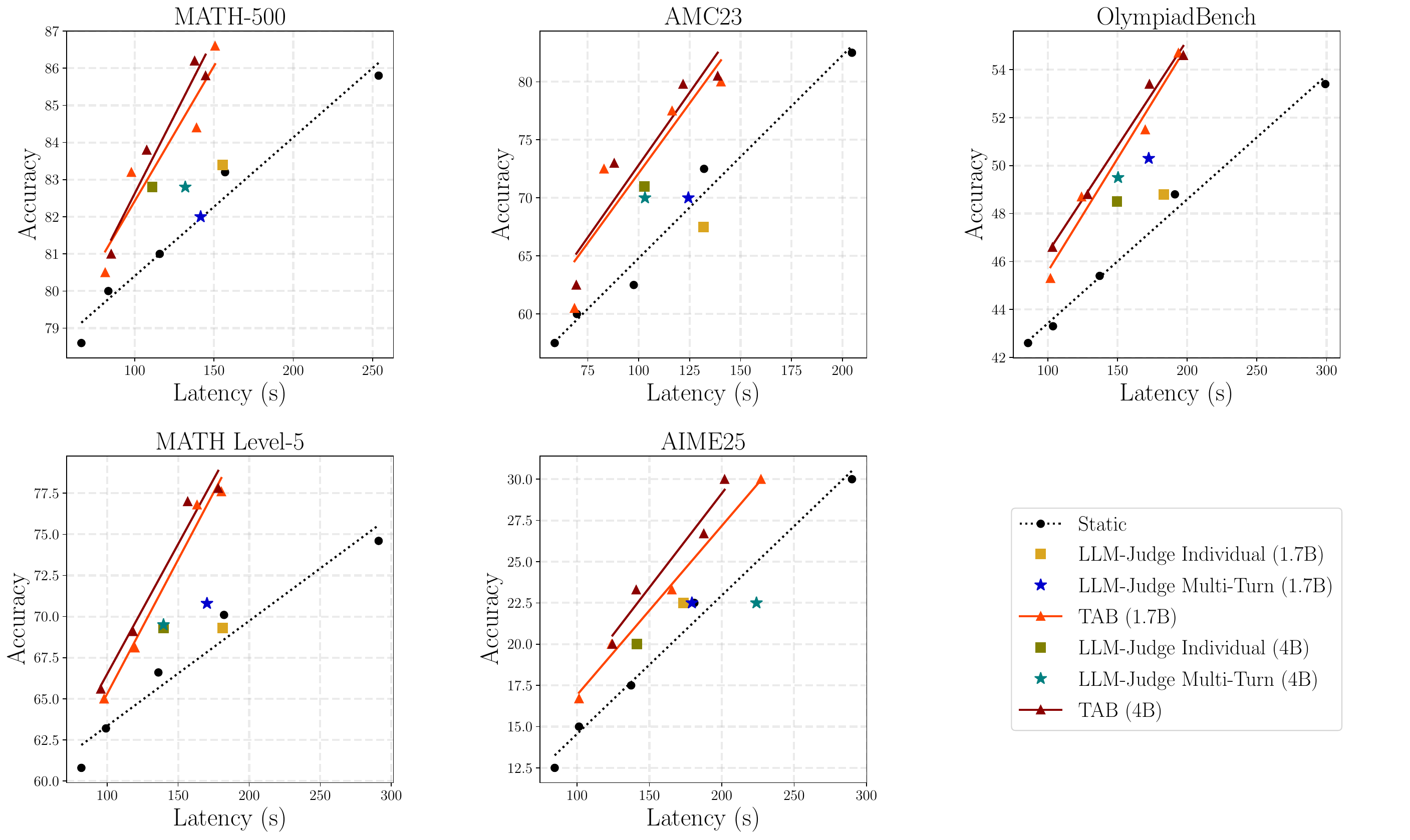}
    \caption{Per-dataset accuracy-latency tradeoff of \tab~on math reasoning tasks, corresponding to the macro average reported in \Cref{fig:latency}. \tab~takes up to 30\% lesser time in solving tasks despite the latency overhead of routing queries to a Budgeter.}
    \label{fig:applatency}
\end{figure}

\paragraph{Training and Serving Setup.} We LoRA finetune \citep{hu2022lora} with rank $r=64$ and $\alpha = 128$ for 125 steps, learning rate $10^{-5}$, batch size $64$ $(8 \times 8)$ and temperature $0.6$, with maximum context length 32k and per-turn responses limited to 4k tokens. A separate Budgeter policy is trained for every global per-problem budget $B \in \{3k, 5k, 8k, 10k\}$. The \tab~All-SubQ policies are likewise trained separately with the augmented state $s_t = [x_{1:t-1}, s'_t]$, $s'_t = [q_1, \dots, q_T]$, and are not obtained by prompting a sequential policy differently at evaluation time. All Qwen models, that is the Budgeters, the User LLM and the L1-Qwen3-8B-Exact Solver, are served locally with vLLM on NVIDIA H100 GPUs, while all GPT-5.4-nano roles are accessed through the API. Latency measurements in \Cref{fig:agenticLatency,fig:latency,fig:applatency} are wall-clock times under this configuration and therefore include API network time for the GPT-5.4-nano roles.
 

 \newpage
\section{Additional Experimental Results} \label{app:additionalExperiments}
 
This appendix collects the results that supplement \Cref{sec:experiments}. \Cref{app:latency} reports the per-dataset latency breakdown behind the macro average of \Cref{fig:latency}, \Cref{app:allsubq} describes the \tab~All-SubQ variant available when a plan of all turns is known apriori, and \Cref{app:ood} reports how the same MATH-trained policy fares on reasoning tasks outside math.
 
\subsection{Per-Dataset Latency on the Auxiliary Task} \label{app:latency}
 
\subsection{\tab~All-SubQ} \label{app:allsubq}
 
A hierarchical plan-and-execute system commits to the plan $q_1, \dots, q_T$ before any budget is allotted, so the future sub-questions are already available at turn $1$ and \tab~as described in \Cref{sec:algo} does not use them. \tab~\textbf{All-SubQ} is the budget allocation policy that allots tokens based on the conversation trajectory thus-far and all sub-questions past and future as $b_t \sim \pi_\phi(\cdot | x_{1:t-1}, q_1, \dots, q_T)$, so that the state representation here is just the set of questions, $s_t' = [q_1, q_2, \dots, q_T]$, along with the trajectory thus-far. The two variants differ only in this input, and we train a separate Budgeter for All-SubQ with the identical recipe of \Cref{sec:expSetup} rather than prompting \tab~differently. We correspondingly compare against an \emph{LLM-Judge Multi-Turn All-SubQ} baseline, the prompted allocator given all sub-questions as input.
 
\begin{figure}[ht]
    \centering
    \includegraphics[width=0.95\linewidth]{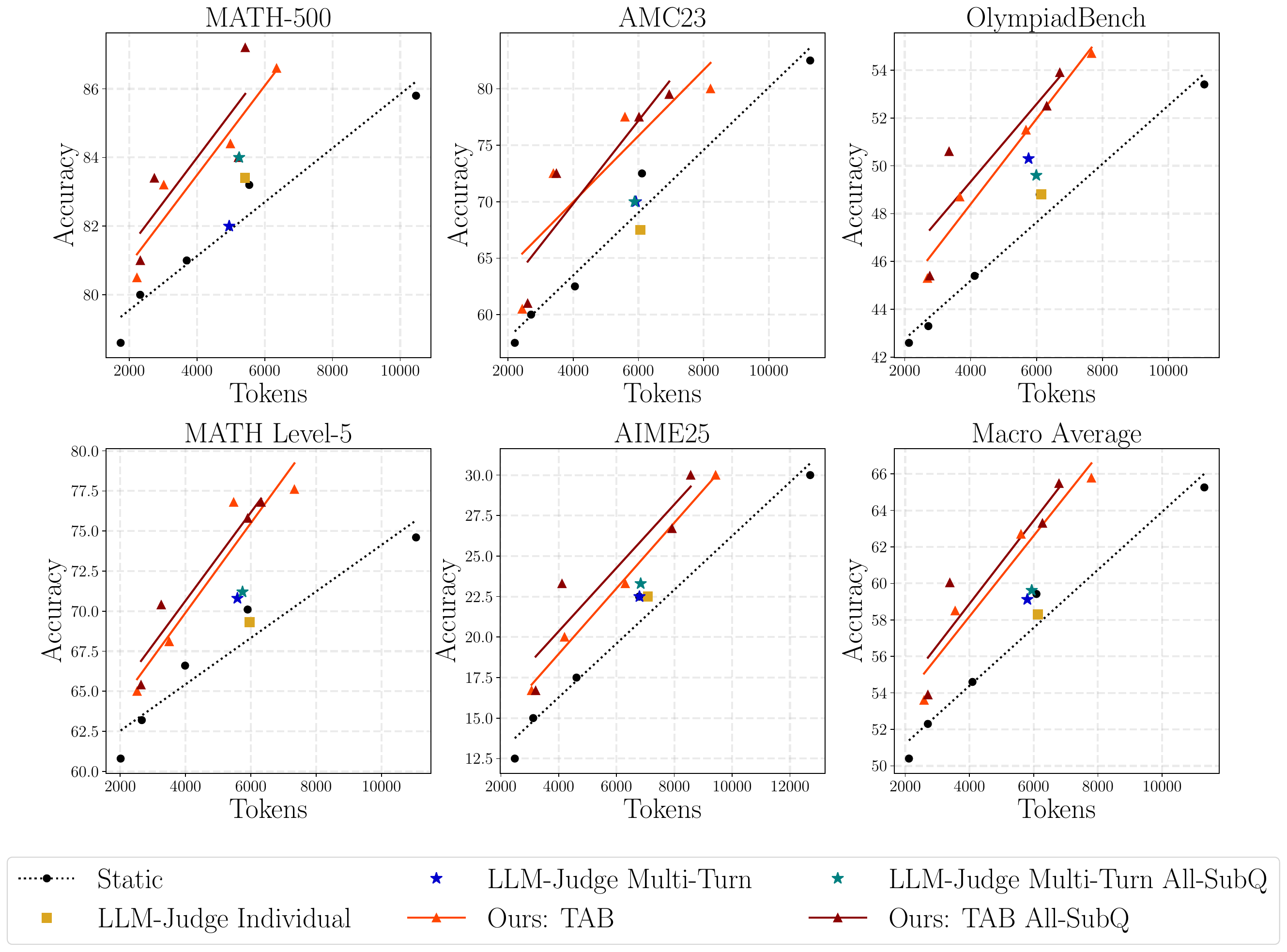}
    \caption{Accuracy-tokens tradeoff of \tab~and \tab~All-SubQ with Qwen3-1.7B budget policies trained with budget penalties $B \in \{3k, 5k, 8k, 10k\}$, across the 5 math reasoning datasets and their macro average. \tab~All-SubQ, given as input all sub-questions, past and future, along with the conversation history, beats a Budgeter without access to future sub-questions. \tab~All-SubQ with $B=10k$ saves $12\%$ tokens over \tab~and up to $40\%$ tokens over baselines.}
    \label{fig:allSubQResults}
\end{figure}

As we observe in \Cref{fig:allSubQResults}, \tab~All-SubQ defines an improved accuracy-tokens tradeoff over \tab, and with $B=10k$ saves $12\%$ tokens over \tab~and up to $40\%$ over baselines without loss in accuracy. Since lookahead is the only difference between the two variants, this gap highlights that explicit knowledge of future steps allows for more precise compute allocation, validating that multi-turn efficiency is fundamentally a global planning problem. Note that this variant is applicable only where a plan of all turns is available apriori, and therefore not in the environment-grounded workflows of \Cref{sec:resultsAgentic}, where the turns are revealed by the environment one at a time.
 
\begin{figure}[hb]
    \centering
    \includegraphics[width=0.95\linewidth]{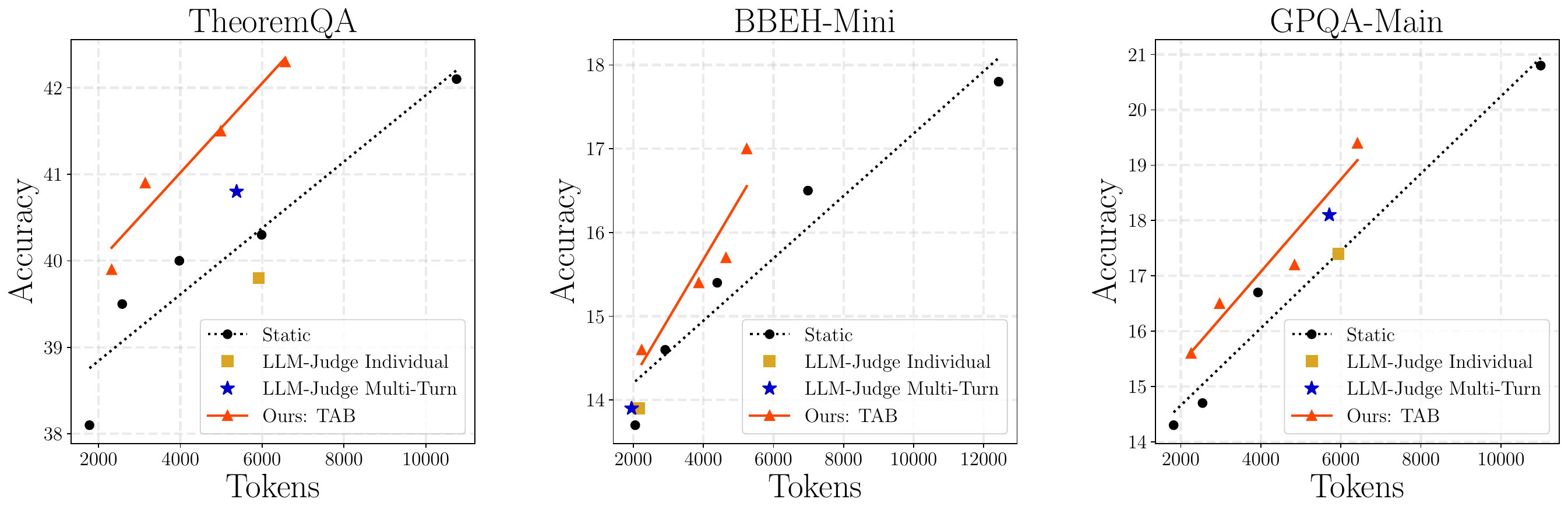}
    \caption{Accuracy-tokens tradeoff of \tab~on reasoning tasks outside the training domain. The budget allocation policy is trained only on Level 5 questions from the train split of MATH and applied here without any further finetuning.}
    \label{fig:ood}
\end{figure}
 
\subsection{Reasoning Tasks Beyond Math} \label{app:ood}
 
\Cref{sec:resultsAgentic} evaluates how the budget allocation policy fares when it is carried from the auxiliary math task to environment-grounded agentic workflows. We additionally report here how it does on multi-turn reasoning tasks that remain hierarchical but are no longer mathematical, using the identical plan-and-execute setup, Solver and MATH-trained Budgeter of \Cref{sec:results}. As shown in \Cref{fig:ood}, \tab~continues to define a superior accuracy-tokens tradeoff on TheoremQA \citep{chen2023theoremqa}, BBEH-Mini \citep{kazemi2025bbeh} and GPQA-Main \citep{rein2024gpqa}, none of which the Budgeter was trained on.

\newpage
\section{Examples of Multi-Turn Reasoning Trajectories in Agents} \label{app:examples}
 
To better understand the strengths of our budget allocation policy \tab~against static and off-the-shelf LLM-Judge baselines, we present four qualitative examples that highlight key characteristics of the learned policy.
 
\subsection{Token Savings via Adaptive Allocation (Tau-Bench)}
This example highlights how \tab~dynamically scales down compute for easy turns, saving tokens without sacrificing accuracy, whereas a static high-effort baseline over-generates redundant reasoning.
 
\begin{figure}[H]
\begin{mdframed}[backgroundcolor=gray!10, linecolor=black, linewidth=1pt]
\textbf{Problem:} 
\footnotesize
\begin{verbatim}
"Hi--my name is Yusuf Rossi (zip 19122). I received order #W2378156,
and I'd like to make an exchange."
\end{verbatim}
\end{mdframed}
\bigskip
 
\begin{mdframed}[backgroundcolor=gray!10, linecolor=black, linewidth=1pt]
\textbf{\tab~Trajectory} (Reasoning Effort: Low):
\footnotesize
\begin{verbatim}
<think>
User provided authentication details. Tool:
find_user_id_by_name_zip. Name: "Yusuf Rossi", Zip: 19122.
</think>
Action Output: [find_user_id_by_name_zip(first_name="Yusuf",
last_name="Rossi", zip="19122")]
 
Result: Success. The agent successfully authenticates using a very
brief reasoning trace, achieving significant compute savings
compared to the static high-effort baseline.
\end{verbatim}
\end{mdframed}
\bigskip
 
\begin{mdframed}[backgroundcolor=gray!10, linecolor=black, linewidth=1pt]
\textbf{Static High Effort Trajectory} (Reasoning Effort: High):
\footnotesize
\begin{verbatim}
<think>
Let's break down the user's request. The user has provided their zip
code (19122) and their name (Yusuf Rossi). My immediate goal is to
authenticate the user so I can access their profile and order
history. The required tool for this is find_user_id_by_name_zip. I
need to ensure both the name and the zip code are passed exactly as
provided. Let's also consider potential edge cases: what if the user
has multiple accounts? What if the zip code doesn't match? I should
prepare to ask clarifying questions if the tool returns a failure.
Once authenticated, I should check the user's recent orders, just in
case they have multiple active issues. Let's call the
find_user_id_by_name_zip tool now... [Continues for 400+ tokens]
</think>
Action Output: [find_user_id_by_name_zip(first_name="Yusuf",
last_name="Rossi", zip="19122")]
 
Result: Success. The agent successfully authenticates, but wastes
compute generating a highly verbose, redundant chain-of-thought for
a deterministic 1-step action.
\end{verbatim}
\end{mdframed}
\end{figure}
 
\newpage
 
\subsection{Adaptive Reasoning Improves Planning (Terminal-Bench)}
This example illustrates how static low reasoning effort fails on complex environment-interaction turns due to an inability to formulate a comprehensive plan, resulting in naive and brittle actions, whereas \tab~correctly scales up the reasoning effort.
 
\begin{figure}[H]
\begin{mdframed}[backgroundcolor=gray!10, linecolor=black, linewidth=1pt]
\textbf{Problem:} 
\footnotesize
\begin{verbatim}
The agent needs to install dependencies and compile SQLite from
source with gcov coverage.
\end{verbatim}
\end{mdframed}
\bigskip
 
\begin{mdframed}[backgroundcolor=gray!10, linecolor=black, linewidth=1pt]
\textbf{\tab~Trajectory} (Reasoning Effort: High):
\footnotesize
\begin{verbatim}
Action Output:
[
{"function_name": "bash_command", "arguments": {"keystrokes":
"apt-get update && apt-get install -y build-essential curl\n"}},
{"function_name": "bash_command", "arguments": {"keystrokes": "curl
-L -o sqlite-autoconf.tar.gz https://sqlite.org/...tar.gz\n"}},
{"function_name": "bash_command", "arguments": {"keystrokes": "tar
-xzf sqlite-autoconf.tar.gz --strip-components=1\n"}},
{"function_name": "bash_command", "arguments": {"keystrokes": "if [
-f configure ]; then CFLAGS=\"-O0 -g -fprofile-arcs
-ftest-coverage\" ./configure; fi\n"}},
{"function_name": "bash_command", "arguments": {"keystrokes": "make
clean || true\n"}},
{"function_name": "bash_command", "arguments": {"keystrokes": "make
-j$(nproc)\n"}},
{"function_name": "bash_command", "arguments": {"keystrokes": "mkdir
-p /usr/local/bin\n"}},
{"function_name": "bash_command", "arguments": {"keystrokes": "if [
-f ./sqlite3 ]; then cp -f ./sqlite3 /usr/local/bin/sqlite3.gcov;
fi\n"}},
{"function_name": "bash_command", "arguments": {"keystrokes": "ln
-sf /usr/local/bin/sqlite3.gcov /usr/local/bin/sqlite3\n"}}
]
\end{verbatim}
\end{mdframed}
\bigskip
 
\begin{mdframed}[backgroundcolor=gray!10, linecolor=black, linewidth=1pt]
\textbf{Static Low Effort Trajectory} (Reasoning Effort: Low):
\footnotesize
\begin{verbatim}
Action Output:
[
{"function_name": "bash_command", "arguments": {"keystrokes":
"apt-get update && apt-get install -y build-essential curl\n"}},
{"function_name": "bash_command", "arguments": {"keystrokes": "curl
-L -o sqlite-autoconf.tar.gz https://sqlite.org/...tar.gz\n"}},
{"function_name": "bash_command", "arguments": {"keystrokes": "tar
-xzf sqlite-autoconf.tar.gz --strip-components=1\n"}},
{"function_name": "bash_command", "arguments": {"keystrokes":
"./configure\n"}},
{"function_name": "bash_command", "arguments": {"keystrokes": "make
&& make install\n"}}
]
 
Result: Failed. Without sufficient reasoning compute to plan how to
apply the coverage instrumentation, it falls back to boilerplate
commands and completely omits the required CFLAGS (-fprofile-arcs
-ftest-coverage).
\end{verbatim}
\end{mdframed}
\end{figure}

\newpage
\subsection{Token Savings via Adaptive Allocation (Math)}
In this example of math reasoning task in the hierarchical plan-and-execute framework, we demonstrate that \tab~spends only 2605 tokens to solve the problem accurately as it correctly identifies the difficulty of a sub-question and allots the right number of tokens compared to the LLM-Judge Multi-Turn baseline policy that ends up (over)-spending 6063 tokens.
 
\begin{figure}[H]
\begin{mdframed}[backgroundcolor=gray!10, linecolor=black, linewidth=1pt]
\textbf{Problem:} 
\footnotesize
\begin{verbatim}
Find the remainder when $(5x + 9)^{611} + (x + 5)^{11} + (x -
1)^{11} + 3x^2 + 1$ is divided by $x + 2.$
\end{verbatim}
\end{mdframed}
\bigskip
 
\begin{mdframed}[backgroundcolor=gray!10, linecolor=black, linewidth=1pt]
\textbf{\tab~Trajectory} (Budget: 256 $|$ Actual: 243 tokens):
\footnotesize
\begin{verbatim}
Subquestion: How can we simplify (x + 5)^11 evaluated at x = -2?
Response: <think>
Okay, let's tackle this. The expression is (x + 5)^11 evaluated at x
= -2. First, substitute x = -2 into the base: (-2 + 5) = 3. So now
it becomes 3^11. Wait, 3^11 is a big number. The simplest form is
just 3^11.
</think>
To simplify $(x + 5)^{11}$ at $x = -2$, substitute $x = -2$:
$$(-2 + 5)^{11} = 3^{11}$$
**Answer:** The simplified value is $\boxed{3^{11}}$.
\end{verbatim}
\end{mdframed}
\bigskip
 
\begin{mdframed}[backgroundcolor=gray!10, linecolor=black, linewidth=1pt]
\textbf{LLM-Judge Multi-Turn Trajectory} (Budget: 1024 $|$ Actual: 1,049 tokens):
\footnotesize
\begin{verbatim}
Subquestion: How can we simplify (x + 5)^11 evaluated at x = -2?
Response: <think>
Okay, let's tackle this sub-question. The user is asking to simplify
(x + 5)^11 when x = -2 ... Let me compute that: -2 + 5 is 3.
Expression simplifies to 3^11.
Wait, but why is this important? Well, in the context of the main
problem, we're evaluating the entire polynomial at x = -2. But let
me double-check... Now, the user might be wondering if there's a way
to compute 3^11 without a calculator...
Let me compute it step by step. 3^1 = 3, 3^2 = 9, 3^3 = 27, 3^4 =
81, 3^5 = 243, 3^6 = 729, 3^7 = 2187, 3^8 = 6561, 3^9 = 19683, 3^10
= 59049, 3^11 = 177147. So, 3^11 equals 177,147. However, in the
context of the main problem, when we add (x + 5)^11 and (x - 1)^11,
which becomes 3^11 + (-3)^11... Since 11 is odd... therefore, they
cancel each other out.
</think>
To simplify the expression $(x + 5)^{11}$... [Continues for another
    200 words] ...
It's worth noting that in the context of the main problem, this term
combines with $(x - 1)^{11}$... leaving no contribution to the final
result. However, this sub-question focuses only on simplifying $(x +
5)^{11}$, not its interaction...
**Final Answer:** $\boxed{3^{11}}$
\end{verbatim}
\end{mdframed}
\end{figure}

\newpage
\subsection{Effect of Early-Stage Budgets on Later Quality (Math)}
 
In another example of math reasoning task in the hierarchical plan-and-execute framework, we demonstrate that the Static baseline with a uniform 512 token allocation for every turn allots too few tokens to a crucial step early in the trajectory leading to compounding errors in subsequent steps and eventually getting the answer wrong even though it spends a comparable number of total tokens of 3198 on the problem. In the step shown below, the baseline overruns its 512-token request and wraps up its reasoning prematurely at 653 tokens, resulting in a wrong algebraic simplification. Note that the Solver adheres to budgets softly rather than being truncated at a hard limit, so the failure here is one of insufficient allotted compute and not of hard truncation. Note that the snippet reproduced here shows only this step, and the erroneous simplification it ends on propagates through the remaining turns of the trajectory, with the 'No valid solution.' output produced at the final synthesis turn which is not shown. On the other hand, our policy \tab~allots the appropriate number of tokens (1024) to the step shown below, simplifies the equation correctly and in the end solves the question using a total of 3548 tokens.

\begin{figure}[H]
\begin{mdframed}[backgroundcolor=gray!10, linecolor=black, linewidth=1pt]
\textbf{Problem:}
\footnotesize
\begin{verbatim}
Solve $\sqrt{x + \sqrt{3x + 6}} + \sqrt{x - \sqrt{3x + 6}} = 6.$
\end{verbatim}
\end{mdframed}
\bigskip
 
\begin{mdframed}[backgroundcolor=gray!10, linecolor=black, linewidth=1pt]
\textbf{Static (512) Baseline Trajectory} (Budget: 512 $|$ Actual: 653 tokens)
\footnotesize
\begin{verbatim}
Subquestion: Square both sides of the equation to eliminate outer
    roots.
Response:
<think>
Let me denote A = sqrt((y^2 + 3y -6)/3) and B = sqrt((y^2 - 3y
-6)/3). So the equation is A + B = 6. Squaring both sides gives (A +
B)^2 = 36...
Now compute A^2 and B^2. A^2 = (y^2 + 3y -6)/3. B^2 = (y^2 - 3y
    -6)/3.
Adding A^2 and B^2: (2y^2 - 12)/3.
... [Overruns its 512-token request and wraps up at 653 tokens] ...
</think>
Expanding the left-hand side:
$$ \frac{y^2 + 3y - 6}{3} + \frac{y^2 - 3y - 6}{3} +
2\sqrt{\frac{(y^2 + 3y - 6)(y^2 - 3y - 6)}{9}} = 36 $$
Thus, the equation becomes:
$$ \frac{2y^2 - 12}{3} + \frac{2}{3}\sqrt{(y^2 - 9)(y^2 - 6)} = 36
    $$
\boxed{\frac{2y^2 - 12}{3} + \frac{2}{3}\sqrt{(y^2 - 9)(y^2 - 6)} =
    36}
\end{verbatim}
\end{mdframed}
\bigskip
 
\begin{mdframed}[backgroundcolor=gray!10, linecolor=black, linewidth=1pt]
\textbf{\tab~Trajectory} (Budget: 1024 $|$ Actual: 865 tokens)
\footnotesize
\begin{verbatim}
Subquestion: Square both sides of the equation to eliminate outer
    roots.
Response:
<think>
... [800+ tokens of rigorous expansion, substitutions, polynomial
    isolation] ...
After squaring both sides, we simplify and derive the equation:
$$ y^4 - 21y^2 + 36 = 9\left(400 - \frac{40}{3}y^2 +
    \frac{1}{9}y^4\right) $$
Simplifying further leads to:
99y^2 - 3564 = 0 \implies y^2 = 36 \implies y = 6
Substituting back, $ y = \sqrt{3x + 6} = 6 \implies 3x + 6 = 36
    \implies x = 10 $.
</think>
Answer: The squared equation simplifies to y^2 = 36, giving x = 10.
\end{verbatim}
\end{mdframed}
\end{figure}

\end{document}